# Automatic identification of the area covered by acorn trees in the dehesa (pastureland) Extremadura of Spain

Benjamín Ojeda-Magaña[1*], Rubén Ruelas[1], Joel Quintanilla-Domínguez[2], Leopoldo Gómez-Barba[3], P. Juan López-Herrera[4], José G. Robledo-Hernández[1], and Ana M. Tarquis[5]

[1] Departamento de Ingeniería de Proyectos, Universidad de Guadalajara, José Guadalupe Zuno No. 48, Zapopan, Post Code 45150, México.

[2] Ingeniería en Telemática, Universidad Politécnica de Juventino Rosas, Hidalgo No. 102 Comunidad de Valencia, Santa Cruz de Juventino Rosas, Post Code 38253, México.

[3] Doctorado en Tecnologías de Información, CUCEA, Universidad de Guadalajara, Zapopan, Post Code 45100, México.

[4] School of Agricultural Engineering, Universidad Politécnica de Madrid, Avda. Complutense s/n, Madrid, Post Code 28040, Spain.

[5] Grupo de sistemas complejos (GSC), ETSIAAB, Universidad Politécnica de Madrid (UPM). Ciudad Universitaria, s/n. Madrid, Post Code 28040, Spain.

Correspondence should be addressed to Benjamín Ojeda-Magaña (benjamin.ojeda@academicos.udg.mx)

## Abstract

The acorn is the fruit of the oak and is an important crop in the Spanish *dehesa extremeña*, especially for the value it provides in the Iberian pig food to obtain the "acorn" certification. For this reason, we want to maximise the production of Iberian pigs with the appropriate weight. Hence the need to know the area covered by the crowns of the acorn trees, to determine the covered wooded area (CWA, from the Spanish *Superficie Arbolada Cubierta* SAC) and thereby estimate the number of Iberian pigs that can be released per hectare, as indicated by the royal decree 4/2014. In this work, we propose the automatic estimation of the CWA, through aerial digital images (orthophotos) of the pastureland of Extremadura, and with this, to offer the possibility of determining the number of Iberian pigs to be released in a specific plot of land. Among the main issues for automatic detection are, first, the correct identification of acorn trees, secondly, correctly discriminating the shades of the acorn trees and, finally, detect the arbuscles (young acorn trees not yet productive, or shrubs that are not oaks). These difficulties represent a real challenge, both for the automatic segmentation process and for manual segmentation. In this work, the proposed method for automatic segmentation is based on the clustering algorithm proposed by Gustafson-Kessel (GK) but the modified version of Babuska (GK-B) and on the use of real orthophotos. The obtained results are promising both in their comparison with the real images and when compared with the images segmented by hand. The whole set of orthophotos used in this work correspond to an approximate area of 142 hectares, and the results are of great interest to producers of certified "acorn" pork.

## Introduction

The Spanish *dehesa* is a geographical area with predominance of an agroforestry system of land use and management, based mainly on the extensive livestock exploitation of a continuous area of grassland and Mediterranean trees, mainly occupied by hardwood species of the genus *quercus*. The action of man intervenes in the *dehesas* for its conservation and persistence and has an average tree cover by exploitation of, at least, 10 trees in production per hectare of the genus *quercus* [1].

The average production of oaks oscillates between 8 and 14 kg of acorns per adult tree, though this is tremendously variable between zones, between years and even between trees. On the other hand, In [2] they cite the daily consumption of acorns by Iberian pigs between 7.1 and 8.4 kg, and from 2 to 6 kg of grass. That is to say, this means that an Iberian pig practically consumes the production of 60 oaks during the at least two months that they are in the *montanera*. That is to say, that for pastures with 50-60 trees/hectare, the stocking density should be 1 pig/hectare.

As of Royal Decree 4/2014 of January 10 in Spain [3], it is necessary to estimate the SAC (from now on we will use SAC instead of CWA), which is the percentage of soil covered by the projection of all treetops of the species of *quercíneas* according to the Geographical Information System of Identification of Agricultural Plots (SIGPAC [4]), in order to calculate the Major Livestock Unit (MLU) according to the denomination of animals whose products are going to be commercialized with the "acorn" label established in said Royal Decree. Thus it was established that the number of pigs to be released in the plots and enclosures depended on the percentage of SAC of *quercus* per hectare in the plots (see Table 1). This is very important since this standard approves the quality that must have the meat (the ham), the shoulder and the acorn-fed Iberian pork loin.

Table 1: Livestock load per hectare according to the percentage of SAC[1].

| Covered wooded area (%) | Maximum allowable livestock load (animals/hectare) |
|---|---|
| Up to 10 | 0.25 |
| Up to 15 | 0.42 |
| Up to 20 | 0.75 |
| Up to 30 | 0.92 |
| Up to 35 | 1.08 |
| Higher than 35 | 1.25 |

The automatic analysis of digital images of crops is a very active and attractive field of research in which artificial vision and *agromotics* converge [5]. Identifying the crown of acorn trees in an orthophoto is not an easy task. First, it must be identified that the detected tree is an acorn and that is, by their size, productive; then the most complicated is to delineate the contour of the tree crown, where it starts (the borderline between soil and tree crown), and where it ends (the borderline between tree crown and its shadow). Figure 1 shows some cases on the difficulties for the correct identification of acorn trees, such as the projection of shadows on the same tree or the ground and whose colouration makes it difficult to discriminate. The structure of the tree also produces several regions, some of which can be associated with shrubs, the hollow in between trees or the proximity of mature and young trees may be interpreted as one.

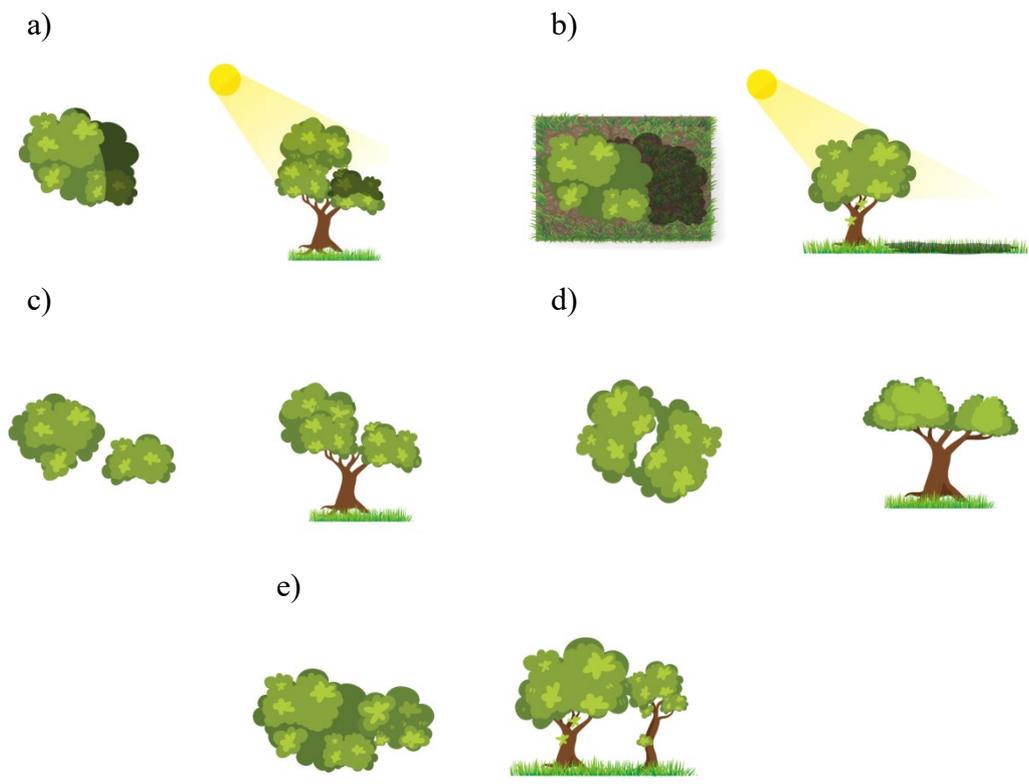

Figure 1: Special cases in the detection of the acorn tree crown area. **a)**Shade on the same tree, **b)**Shade on grass, **c)**Irregular tree, **d)**Hollow and **e)**Trees together.

The segmentation process consists of dividing an image and isolating the regions of interest or discrete objects present in an image or scene, such as the crown of the acorn trees in this study, for a later analysis that can be qualitative or quantitative. The identification of a tree crown or a shrub in an orthophoto is a difficult task because the human being has to differentiate the regions that present at least a certain homogeneity in which some characteristics of the acorn trees must be recognized, such as colour, texture, size and shape.

There are a variety of methods for the process of segmentation of images. In general, these algorithms are divided into the following: grouping and classification [6,7,8] of edge or contour detection [9,10], of growth regions [11], and morphological [12].
The partitional clustering algorithms of the c-Means family [13] divide the colour image into a cartesian RGB feature space in a given number of clusters where each one found is as homogeneous as possible, accordingly with the used optimization criteria.

The detection of plant mass, especially tree canopy, using aerial images (LIDAR images) has been of interest because of the possibility of characterizing forest areas [21] and deriving useful information to calculate approximations of biomass or wildlife habitat [22]. However, the automatic detection, segmentation of treetops involves several challenges, such as the intensity of both color and luminosity, the structural characteristics of both the forest area and the trees and shrubs [23] (when it is possible to have them).

Therefore, in this work we make use of orthophotos obtained from the SIGPAC, and in them we identify the areas that represent the tree crowns and the oak shrubs, as well as what it represents the soil or pastureland. In this study we also use images segmented manual that serve as a comparison element, and an automatic procedure based on the segmentation of images with the GK partitional clustering algorithm (Gustafson Kessel) [14], which was improved by Babushka *and others* [15], and in this study we named this algorithm as GK-B (Gustafson Kessel-Babuska). Later we identify the crown area of the acorn trees, and calculate the percentage of SAC.

The rest of this document is organised as follows; Section 2 describes the proposed method, the manual segmentation whose results are used as comparison elements, as well as the set of images used in the experiment. The experimental results are given in Section 3. The analysis and discussion of SAC detection are given in Section 4, while the main conclusions of this work are presented in Section 5.

## Materials and Methods

### DataSet

The orthophotos used in this study are from a holm oak *dehesa* in the town of Alconchel, province of Badajoz in Spain, near a pond that forms the Bufanda stream, and are from an area between the Nateras farmhouse and the Barrancón farmhouse, located at the geographical coordinates with Latitude 38º 35'N and Longitude 7º 14'W with an altitude of around 180 m above sea level. The zone of the images is very close to the Guadiana River that borders Portugal.

The orthophotos used in this study have been taken from the Geographic Information System of Agricultural Plots (SIGPAC), which main purpose is to geographically identify the parcels of farmers and ranchers, lands related to cultivated areas and those used by livestock. The digital orthophotos of this study were provided by the National Geographic Institute, belonging to the Ministry of Development, and the flight is from 2011, with pixel size 0.25m, in BMP (bitmap) format with the corresponding TFW file for georeferencing.

For this work, 38 aerial orthophotos (Im_58 to Im_95) were used in three color bands (RGB), from the area between the Nateras farmhouse and the Barrancón farmhouse (in Spain), sizes did vary by a few pixels, though at the end they were cut so that they were all homogeneous and had a size of 256x256 pixels. The orthophotos of this case study are consecutive images in horizontal form. See example in Figure 2.

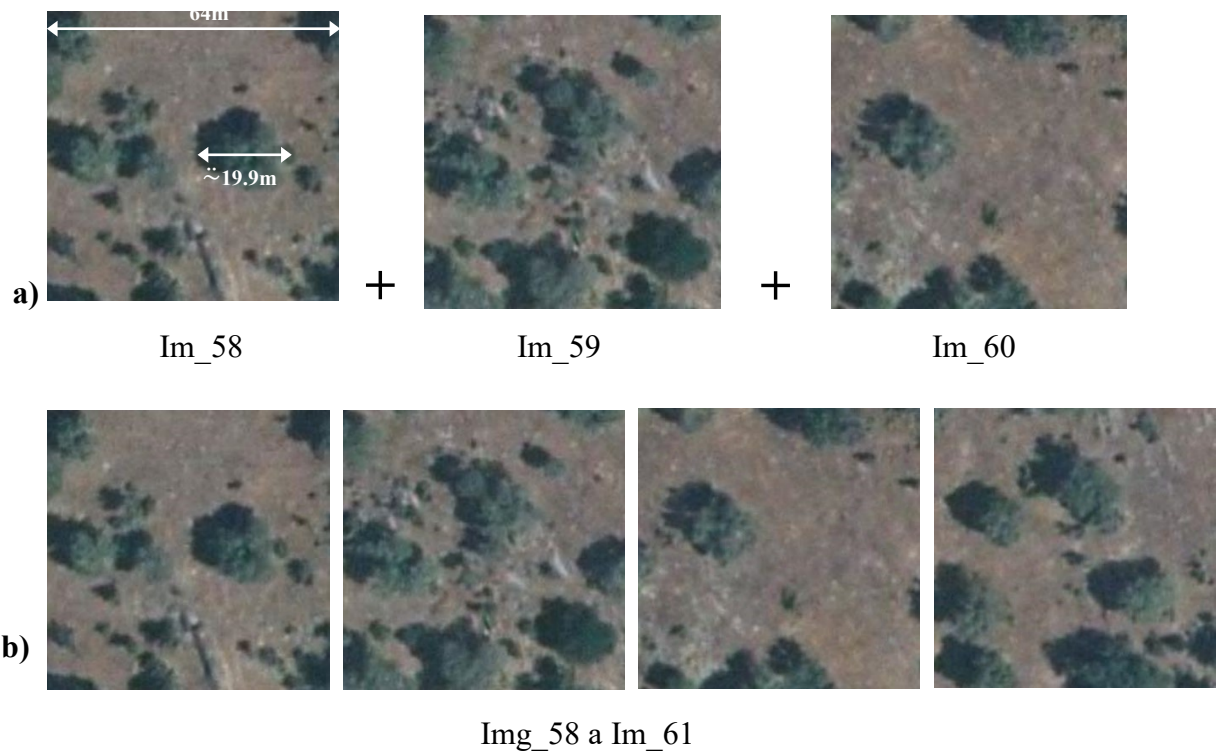

Img_58 a Im_61

Figure 2: Original aerial orthophotos of the *dehesa* (Nateras) of 256 x 256 pixels. **a)** Individual images, **b)** Reconstruction with consecutive images.

In the 38 orthophotos studied, each pixel represents $0.25m$ (pixel size), so each image represents $4096m^2$ in area corresponding to 64m x 64m. Thus, in this work we analyze 38 images aligned. Hence we have $64m$ x $2432m$ which represents approximately 15.56 hectares of the dehesa extremeña.

**Detection of the crown of the acorn trees**

In this study we work to identify three regions of interest or classes in each of the 38 digital aerial orthophotos and corresponding to i) surface of the acorn tree crown, ii) crown of acorn shrub and iii) soil. The first region (i) consists of the crown of oak trees (*quercus*) that may be producing acorns, which is the one of the most interest, to later calculate the percentage of SAC, the second region (ii) is the crown of the young shrubs that still do not produce acorns (shrubs) or shrubs that are not acorn, while the last class (iii) refers to the soil (pastureland and/or stones).

**Manual Segmentation**

The manual segmentation of the 38 orthophotos has been carried out by experts, who have visually identified the regions of interest through computer images. Thus, each of the 38 orthophotos has been delineated, by contour, the region of interest differentiating between the crowns of the acorn trees and the shrubs. Figure 3a shows the results of said manual segmentation in the first four consecutive images (Im_58 to Im_61). Here you can see the contour is drawn in green that identifies the crown of the acorn trees, while the contour in yellow represents the crown of the shrubs; the rest of each image corresponds to the soil, pastureland or stones. Once the contours of the crown of the trees and the acorns were identified, then proceeded to fill in the different areas and in this way have a clearer idea of the SAC in each of the orthophotos (see Figure 3b).

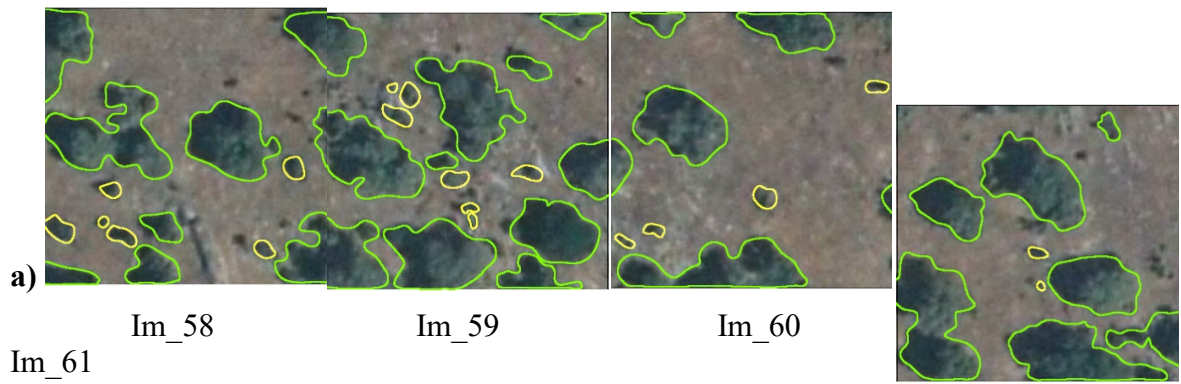

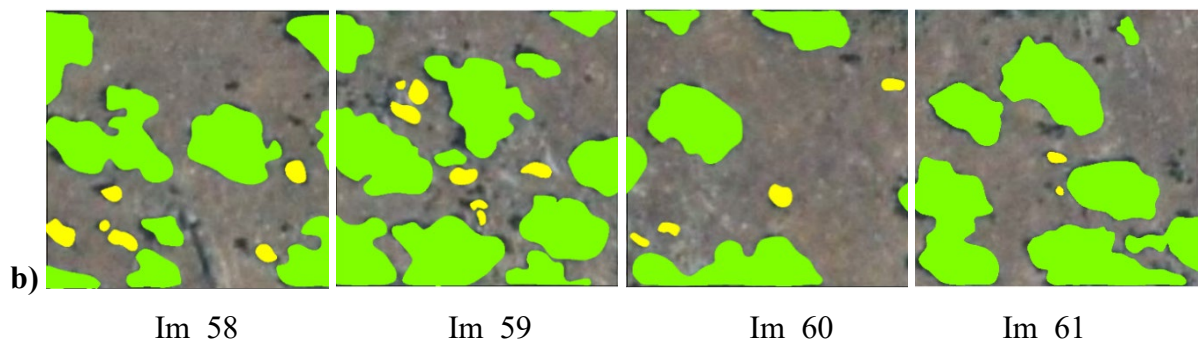

Figure 3: Segmentation of the orthophotos Im_58 to Im_61. **a)** Detection of the contour of the crown tree (green) and acorn shrubs (yellow), **b)** Manual segmentation of the acorn trees (green) and acorn shrubs (yellow).

In the first orthophotos (Figure 3a, Images Im_58 - Im_61), the contrast between the soil and the crown of the trees and shrubs is very high, so its detection does not represent a problem. Also, the trees are separated from each other, and this also facilitates their identification. However, in other orthophotos (with crowded trees/shrubs), both the contrast and the separation do not favour detection, in addition to the shadows that can be overlapping on trees or shrubs. But conveniently we use the results of the manual segmentation as comparison element against the automatic segmentation.

The following procedure was applied to each of the 38 orthophotos:

Manual Segmentation

  i. Identification of acorn trees, and delimitation of the contour of the trees crown.

  ii. Identification of the acorn shrubs, and delimitation of the contour of the crown of the shrubs.

  iii. Binarization of manually segmented images.

     a. Obtaining binary images of acorn trees.

     b. Obtaining the binary images of the acorn shrubs.

  iv. Estimation of SAC percentage for acorn trees.

v.  Calculation of the homogeneity (value of measure NU - Non Uniformity) of the regions of the three elements to be identified: trees, shrubs and soil.

Figure 4 shows the results of binarizing the orthophotos that have been manually segmented. In Figure 4a the white pixels represent the acorn tree crowns and the acorn shrubs, while the black pixels represent the ground. Figures 4b and 4c show binary images of acorn trees and acorn shrubs, respectively. Thus, once you have the binary images with only the crown of the acorn trees, you can estimate the SAC value (%).

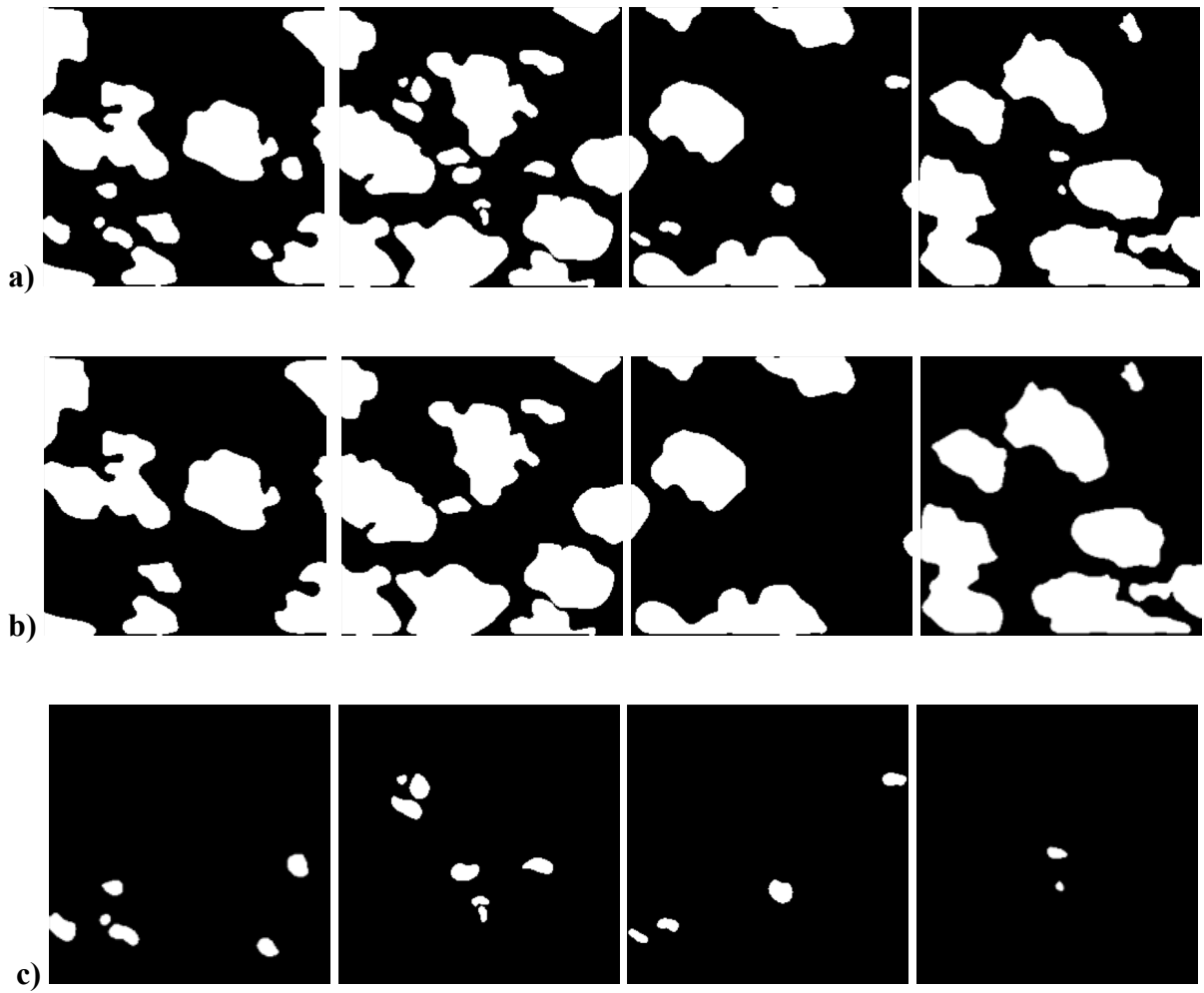

Figure 4: The result of the binarization of the manual segmentation (Im_58 to Im_61). **a)** Binarization of the crown of the acorn trees (white), acorn shrubs (white) and soil (black), **b)** binary images of acorn trees (white), **c)** binary images of acorn shrubs (white).

**Automatic segmentation by the GK-B algorithm**

For this case study of automatic segmentation, a partitional clustering algorithm has been used, derived from the c-Means family, corresponding to the Gustafson Kessel algorithm (GK) [14]. One of its main advantages is that it identifies groups with a hyper-ellipsoidal shape in the space of the characteristics, due to it is based on the Mahalanobis distance instead of the Euclidean distance used by other algorithms such as the Fuzzy c-Means [16]. For this case study, the space of the characteristics corresponds to the orthogonal RGB space because we only use the color, so that the coloured regions form hyper-ellipsoidal shape groups [17].

Babuska *and others*. [15] made an improvement to the GK algorithm so that the groups adapt better to their natural form and it is this algorithm, which we call here GK-B, which we have used for the process of automatic segmentation of the orthophotos. This algorithm requires the adjustment of the parameters $\rho_i$ (cluster volumes), $\beta$ (condition number threshold) and $\gamma$ (weighting parameter); regularly the first two parameters remain constant $\rho_i = 1, \ldots, c$, and $\beta = 10^{15}$, the parameter that changes is $\gamma$ which takes values in $(0,1)$. Besides, there are also the standard parameters as $m$, related to the fuzziness of the partition, and $c$, the number of clusters or objects to find in an image.

In this proposed approach the initial number of classes to be identified is two, since the tree crowns and the acorn shrubs are very similar, and therefore they are identified in a single class, the other class corresponds to the ground soil. When a good identification is not achieved, the number of classes is increased to 3, this has happened when there are just few acorn trees in the image and most of the image is soil, eg. Figure 5 shows few acorn trees and its corresponding area is relatively small against the size of the whole image. Thus, if the detection is not good enough (by visual inspection), the number of classes is increased and the image is processed for segmentation again.

**Clustering algorithm GK-B** [14] [15]

Given the dataset $Z = \{z_{n1}, z_{n2}, \ldots z_{nN}\}$, where $n$ represents the number of features (color layers in RGB) and $N$ the number of pixels of the image, choose an integer number of clusters $c \in \{1, 2, \ldots, N\}$, c < N, the weighting exponent $m > 1$ and the termination tolerance $\varepsilon > 0$.

_______________________________________________________________________________

1. Initialize randomly the membership matrix $U_{inicial}$ with $\mu_{ik} \in \{0,1\}$ and $\sum_{k=1}^{N} \mu_{ik} = 1$.
2. Calculation of the prototypes of the clusters according to the number of data $N$ and the number of classes $c$.

$$v_i = \frac{\sum_{k=1}^{N}(\mu_{ik})^m z_k}{\sum_{k=1}^{N}(\mu_{ik})^m}, 1 \leq i \leq c. \tag{1}$$

3. Calculation of the las covariance matrices for the clusterings,

$$F_i = \frac{\sum_{k=1}^{N}(\mu_{ik})^m (z_k - v_i)(z_k - v_i)^T}{\sum_{k=1}^{N}(\mu_{ik})^m}, 1 \leq i \leq c. \tag{2}$$

   a. Estimation of the covariance matrix for the GK-B algorithm

$$F_i = (1 - \gamma)F_i + \gamma \, det(F_O)^{\frac{1}{n}} I, \tag{3}$$

   b. Extract eigenvalues $\lambda_{ij}$ and eigenvectors $\emptyset_{ij}$ from $F_i$.
   c. Find $\lambda_{ijmax} = max_j \lambda_{ij}$ and set:

$$\lambda_{ij} = \lambda_{imax}/\beta \quad \forall_j \text{ such that } \lambda_{imax}/\lambda_{ij} > \beta. \tag{4}$$

   d. Reconstruction of the covariance matrix $F_i$ by

$$F_i = [\emptyset_{i1\ldots}\emptyset_{in}] diag(\lambda_{i1,\ldots}\lambda_{in,\ldots})[\emptyset_{i1\ldots}\emptyset_{in}]^{-1} \tag{5}$$

4. Calculation of the distances:

$$D_{ik}^2 = (z_k - v_i)^T \left[ \rho_i \, det(F_i)^{\frac{1}{n}} F_i^{-1} \right] (z_k - v_i), 1 \leq i \leq c, 1 \leq k \leq N. \tag{6}$$

5. Actualization of the membership matrix if $D_{ik} > 0$

$$U = \frac{1}{\sum_{j=1}^{c} (\frac{D_{ik}^2}{D_{jk}^2})^{\frac{2}{(m-1)}}}, 1 \leq i \leq c, \ 1 \leq k \leq N, \tag{7}$$

6. Verification of whether the last variation of the membership matrix exceeds a value $\varepsilon$, small and previously established. If so, go back to step 3, else stop.

$$\|U_{new} - U_{old}\| < \varepsilon \tag{8}$$

---

Proposed methodology for automatic segmentation of orthophotos.

---

i. Initial segmentation of the 38 orthophotos using the GK-B algorithm and the following parameters: $\rho_i = 1, ..., c$, $\beta = 10^{15}$, $\gamma = 0$ y $m = 2$, y $\varepsilon = 10^{-3}$ for all the orthophotos. The initial number of clusters is $c = 2$.

ii. Visual inspection and evaluation of the quality in the crown tree detection, if the result is not satisfactory enough then increase, by one, the number of classes.

iii. Binarization of the segmented images in two regions: $R_1$ for the crown tree area and shrubs (white pixels) and $R_2$ or the soil(black pixels).

iv. Separation of the region $R_1$ in tree crown ($R_{1a}$) and acorn shrubs ($R_{1b}$) according to the following rule: if there is an isolated blob less than or equal to an area of 1,650 pixels, separate it and label it as a shrub, in such a way that:

$$R_1 = R_{1a} \cup R_{1b} \tag{9}$$

v. Once that the blobs exclusive of the acorn trees have been identified ($R_{1a}$), proceed to estimate the percentage of the covered wooded area (SAC).

vi. Calculate the value of NU for regions $R_{1a}, R_{1b}$ y $R_2$.

---

**Performance measures**

Given the availability of manually segmented images, these are compared against the images segmented with the GK-B algorithm. The ratios for false positives (*FPR*) and false negatives (*FNR*) are used for comparison, knowing that the FPR values correspond to the ratio of the number of pixels in the background ($B_0$, floor of the manual segmentation) misclassified as Foreground ($F_T$, acorn tree crown of the automatic segmentation) to the total number of background pixels in the image segmented by hand. Here, the symbol $|A|$ represents the cardinality of set $A$. To calculate this parameter, we use Equation 10.

$$FPR = \frac{|B_0 \cap F_T|}{|B_0|} \tag{10}$$

On the other hand, the *FNR* ratio corresponds to the Foreground number ($F_O$, acorn tree crown of the manual segmentation), pixels misclassified into Background ($B_T$, solid space of the automatic segmentation) to the total number of foreground pixels in the ground truth image; and it is calculated with Equation 11.

$$FNR = \frac{|F_o \cap B_T|}{|F_o|} \tag{11}$$

The values *FPR* and *FNR* are in the interval [0,1]. On the other hand, values close to zero are indicative of high similarity between the results of manual segmentation and automatic segmentation, while values close to one are indicative of excessive over-segmentation and under-segmentation, respectively.

To analyze the similarity between the results of manual segmentation *vs* automatic segmentation, the Jaccard's similarity index (*ISJ*) [18] is used, which is a statistical method to compare similarity and diversity between data sets, defined as the size of the intersection divided by the size of the union of the data sets, as shown in Equation 12:

$$ISJ = \frac{|I_{SA} \cap I_{SM}|}{|I_{SA} \cup I_{SM}|} \tag{12}$$

Where $I_{SA}$ represents the binary image obtained from the automatic segmentation and $I_{SM}$ represents the binary image obtained from the manual segmentation. If the ISJ values are near to 1 ensure that segmentation is closer to the reference standard.

On the other hand, the value of a measure of non-uniformity (*NU*) provides an homogeneity degree of each of the segmented regions in the interval (0,1), and helps to better understand the process of separation of objects in the images, when "ground truth" information is not available [19, 20]. This measure is calculated with Equation 13,

$$NU = \frac{|F_T|}{|F_T + B_T|} \cdot \frac{\sigma_P^2}{\sigma^2} \tag{13}$$

Where $|F_T|$ represents the number of pixels in the region of the crown of the acorn trees, $|F_T + B_T|$ the total number of pixels in the image, $\sigma_P^2$ the variance of the pixels, single-band RGB color space, this can be R and, by last, $\sigma^2$ corresponds to the total variance of the intensity values of the pixeles of the color band in question. The previous process is repeated for the other two bands (GB), and subsequently, a NU average is obtained for each of the images. The calculated NU values for a well-segmented RGB image take values close to zero; whereas the worst segmentation cases take values close to one.

**Comparison criteria**

Humans detect the crown of trees and acorns from orthophoto through their colour and homogeneity, that is, the pixels that have a green colour or close to it and are similar to each other, this process is carried out subjectively. The differentiation between acorn trees and shrubs is done only by its size. Experts determine the contour of the tree crown by discriminating pixels of trees on the environment, including the shadows.

The degree of homogeneity of the objects of interest found with manual segmentation and with automatic segmentation is estimated with a NU value, as described in the materials and methods section. A value close to zero indicates that the intensity of the RGB pixels of the tree crown identified is very homogeneous. Therefore, the value of NU is calculated for the region of interest corresponding to the acorn tree crown, both for manual segmentation and for automatic segmentation, and thus assess the segmentation process.

## Results and Discussion

**Identification of the crown of the acorn trees and shrubs using the GK-B algorithm**

Of the 38 orthophotos used in this study, 34 were segmented into two classes and only four into three classes with the GK-B clustering algorithm, because these four orthophotos contained very few acorn trees, and there are a large number of pixels of soil (see Figure 5). Therefore, in these cases, the soil is segmented into two classes, while the other class identifies the acorn trees and shrubs. Figure 6 shows the orthophotos Im_58 to Im_61 segmented into two classes: acorn trees and shrubs, on the one hand, and soil and stones on the other. The values used for the parameters were: $\rho_i = 1, \ldots, c$, $\beta = 10^{15}$, $\gamma = 0$, $m = 2$ and $\varepsilon = 10^{-3}$ for all orthophotos. The initial number of clusters was $c = 2$. In the orthophotos of Figure 6, each class is represented with the average value of the intensity of the pixels identified in the RGB space. In this case, the green colour represents the crown of the trees and the acorns, while the brown colour represents the soil. The difference in tonalities between the images is due to the difference in soil homogeneity.

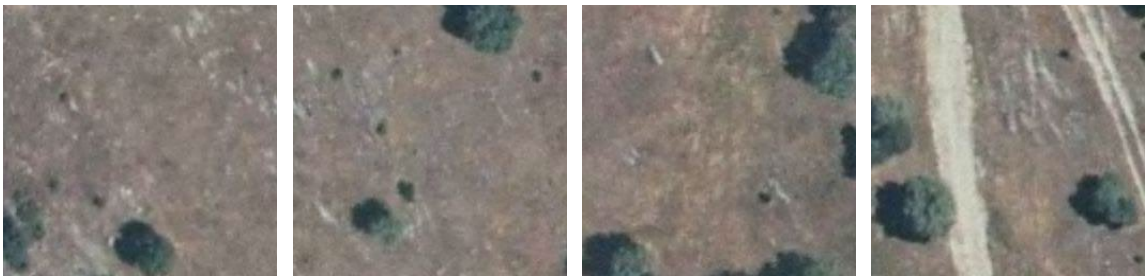

Figure 5: Orthophotos Im_62, Im_63, Im_64 and Im_67 were segmented into three classes.

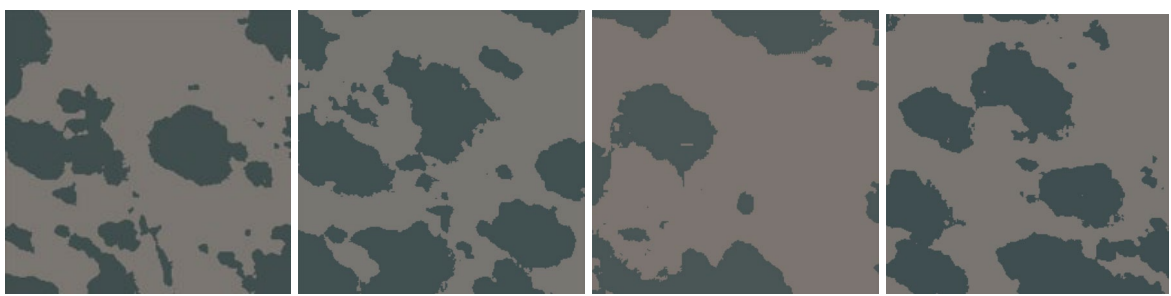

Figure 6: Segmentation of the orthophotos Im_58 to Im_61 with the GK-B algorithm and representation with the average RGB value of each segmented class.

Figure 7a shows the segmentation with the original pixels of the soil where green pixels represent the crown area of the acorn trees and shrubs. Figure 7b shows the separation of acorn trees and shrubs, representing the latter in yellow.

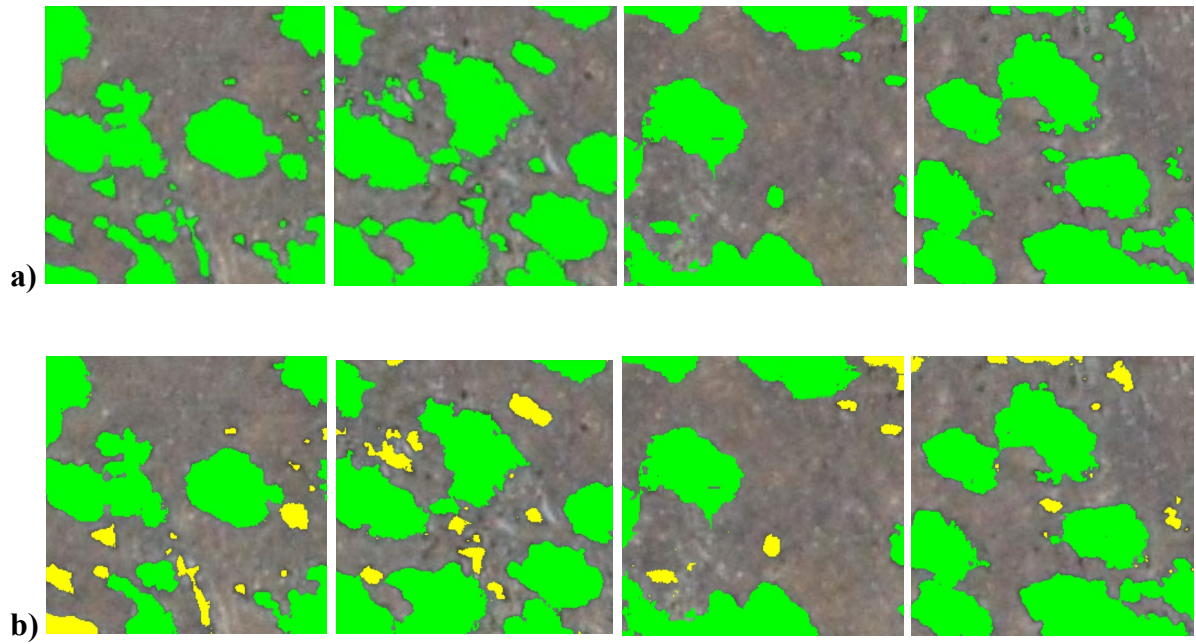

Figure 7: Representation of the automatic segmentation of orthophotos Im_58 to Im_61. **a)** acorns trees and shrubs (green), **b)** acorn trees (green) and acorn shrubs (yellow).

As with manual segmentation, here also a differentiation is made between the acorn tree crown and shrubs. However, in this work, the difference depends on the size of the identified blob. Thus, if a blob is higher than a threshold of 1,650 pixels it is considered as a tree crown; otherwise, is considered as a shrub.

To determine the percentage of SAC in the segmented orthophotos first is necessary to binarize them. Thus, in Figure 8a shows the binary orthophotos, where the black colour represents the soil and white colour the acorn tree crown and shrubs. Figure 8b shows only the area of the crown of the acorn trees, while in Figure 8c only the acorn shrubs are shown.

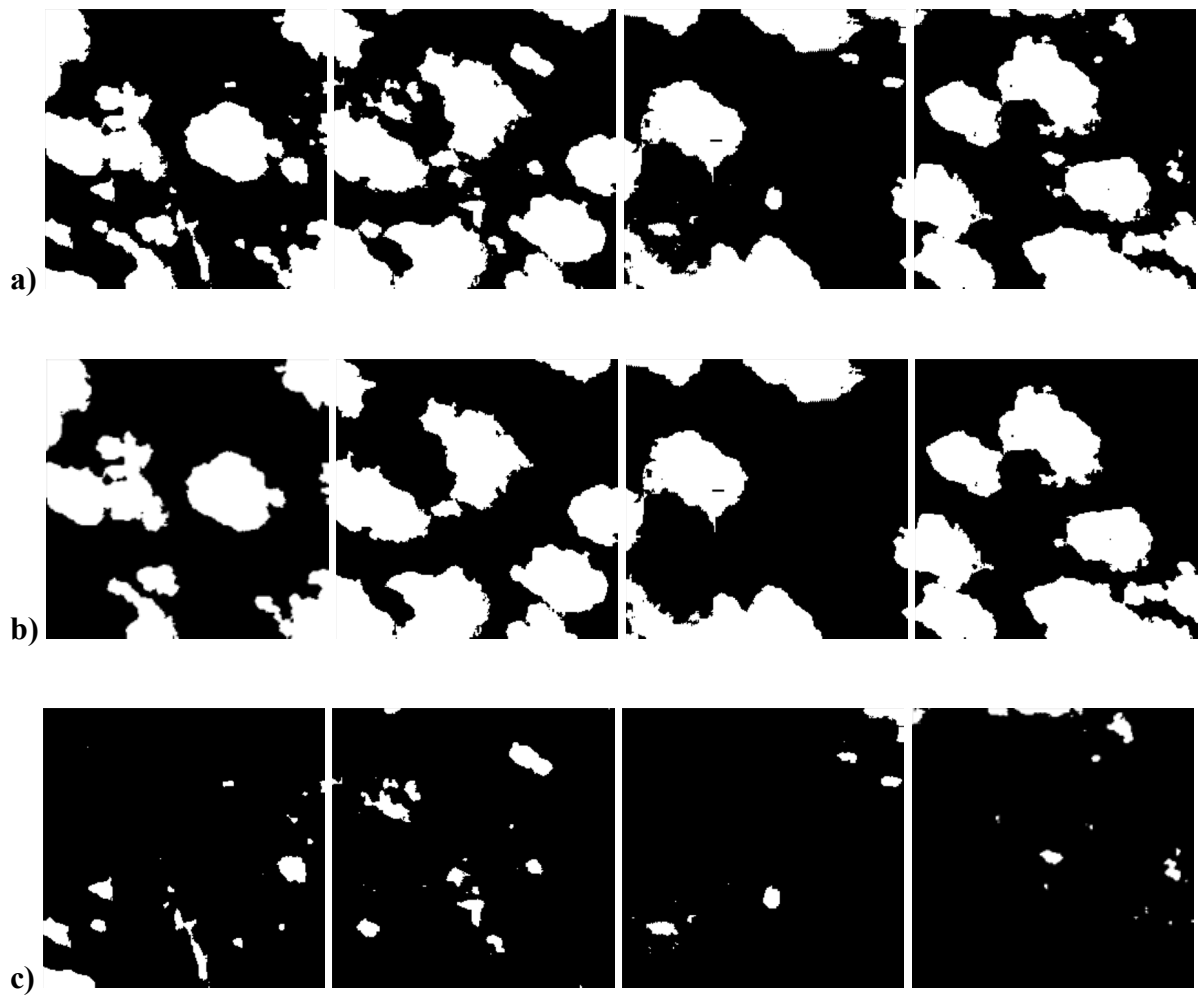

Figure 8: Orthophotos Im_58 to Im_61 segmented with the GK-B algorithm and binarized. **a)**Crown of trees and shrubs in white colour, **b)**Acorn trees, **c)**Acorn shrubs.

**Jaccard's Index of Similarity**

Figure 9 shows the values of the *ISJ* for the comparison of the results of automatic segmentation against manual segmentation. In the first 14 orthophotos, the highest *ISJ* values were obtained between 0.83 and 0.7.

Figure 10a shows the original orthophoto (to the left) and the segmentations that have given the highest similarity. However, from Image 15b, which is when the experts begin to discriminate the shadows, the *ISJ* is reduced indicating more significant differences between the segmentation. From orthophoto 21 (similar to Figure 10b), where the trees are very crowded, the *ISJ* falls to a relatively low value related to the rest of the comparisons. See Figure 10b where this orthophoto appears, as well as its manual and automatic segmentation.

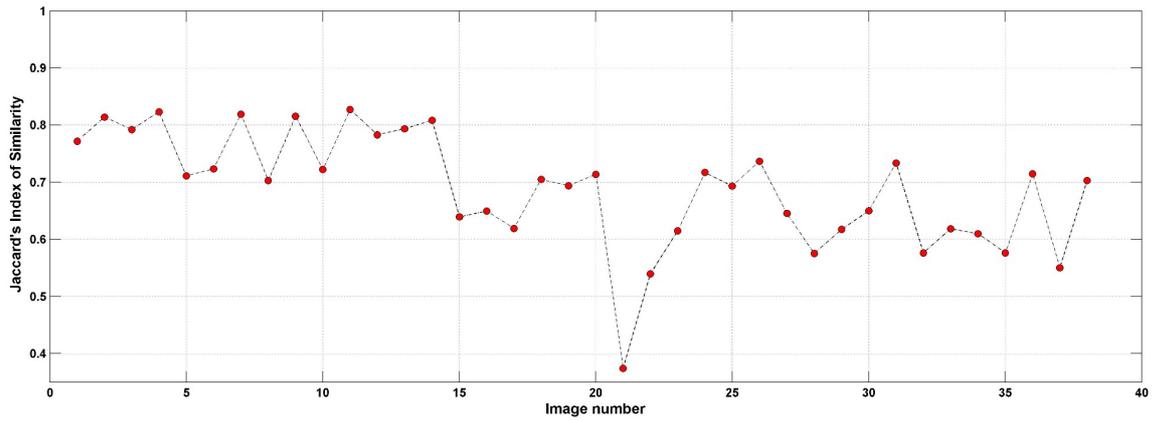

Figure 9: Values of the Jaccard similarity index for the 38 orthophotos.

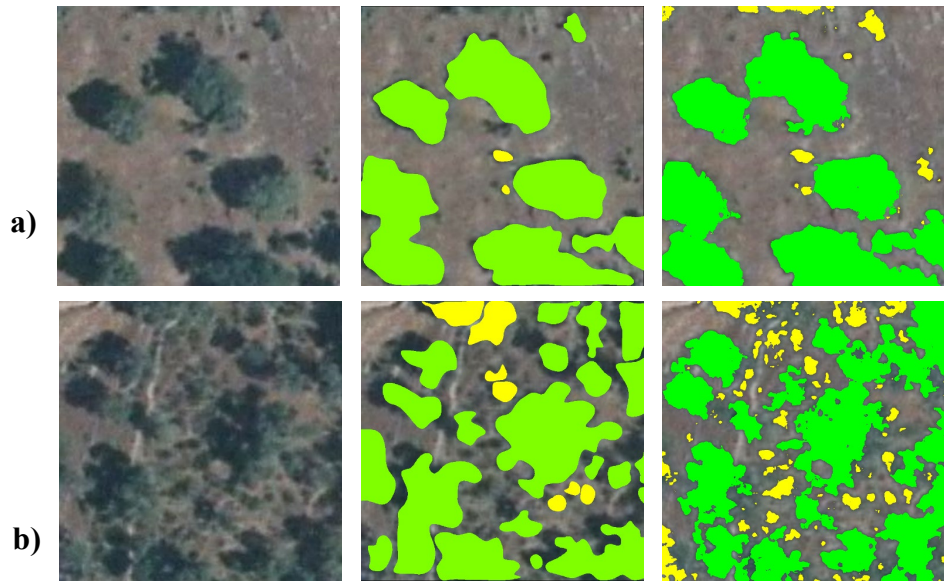

Figure 10: Better and worse value according to the ISJ. **a)** orthophoto Im_61 original, manual segmentation and automatic segmentation, **b)** orthophoto Im_78 original, manual segmentation and automatic segmentation.

**Homogeneity in acorn trees and shrubs**

Figure 11 shows the NU values of the 38 orthophotos for manual segmentation (red dots) and automatic segmentation (green dots). Here it is observed that, except for six cases, the value of NU is smaller for the first 24 orthophotos with automatic segmentation, while for the last 14 orthophotos, except for orthophoto 31, the value of NU is smaller for the manual segmentation. However, we observed a correlation between the NU values of both segmentations for most of the orthophotos, which is also an indication of the difficulty for very accurate identification of the crown of the acorn trees and shrubs.

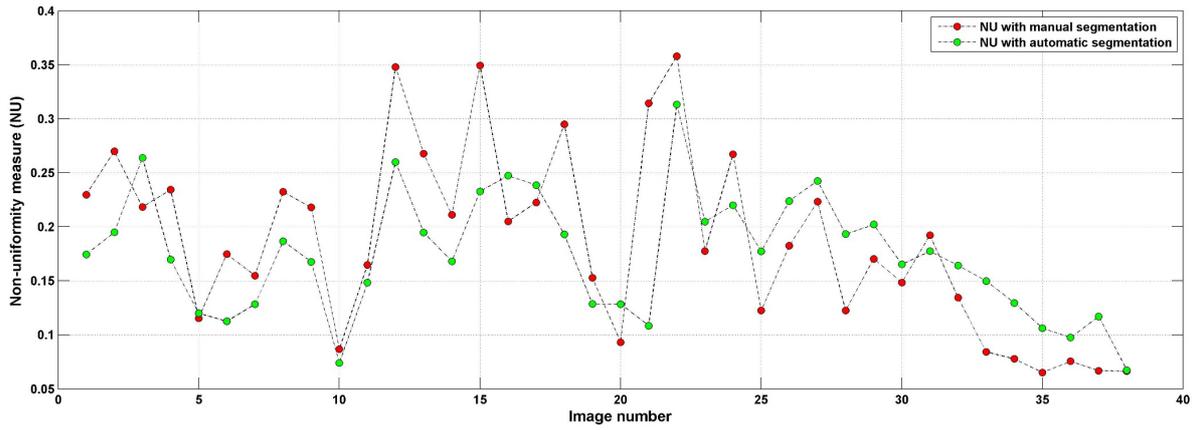

Figure 11: NU value of each of the 38 orthophotos for manual and automatic segmentation.

## SAC percentage

From the segmentation results of the 38 orthophotos (Im_58 to Im_95), the calculation of the area covered in each of the orthophotos was made, see Figure 12, and then an average was calculated, considering the trees separately. Acorn, shrubs and soil. The results obtained, both for manual segmentation and automatic segmentation, are shown in Table 2, in addition to the corresponding average NU values.

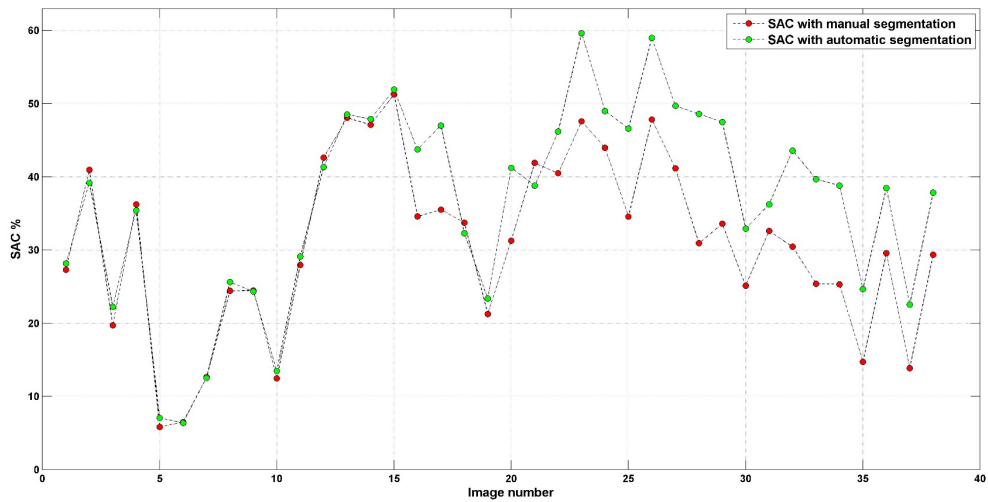

Figure 12: SAC percentage for the manual y automatic segmentation.

Table 2: Average SAC and NU values of the manual y automatic segmentations.

|  | Manual segmentation | | Automatic segmentation | |
| --- | --- | --- | --- | --- |
|  | SAC (%) | NU | SAC (%) | NU |
| Acorn trees | 30.85 | 0.186 | 36.07 | 0.173 |
| Shrubs | 01.23 | 0.004 | 01.80 | 0.007 |
| Soil | 67.92 | 0.481 | 62.13 | 0.224 |

**Comparison between manual segmentation and automatic segmentation**

In Figure 13a the orthophotos Im_70 to Im_73 (corresponding to images 13 to 16) of the study area are observed, while Figure 13b shows the results of the manual segmentation and in Figure 13c the results of the Automatic segmentation concerning the original orthophotos in both cases and in Figure 13d illustrates the comparative results between automatic segmentation and manual segmentation. Thus, taking the latter as a reference, there are the false positives in blue and the false negatives in orange, that is, in blue the pixels that the automatic segmentation detects in more respect to the manual segmentation and orange the pixels that it detects of less, also concerning the manual segmentation.

The value of NU is smaller for automatic segmentation than for manual segmentation (See Table 2), because the pixels of the shadow have high similarity in colour to the pixels of the acorn trees, and in the automatic segmentation they are recognised as of the same class. A different case occurs with manual segmentation where, although some of the shadows are no longer detected, the contours of the acorn trees are not well identified, because they also take up soil pixels, causing an increase in the NU value.

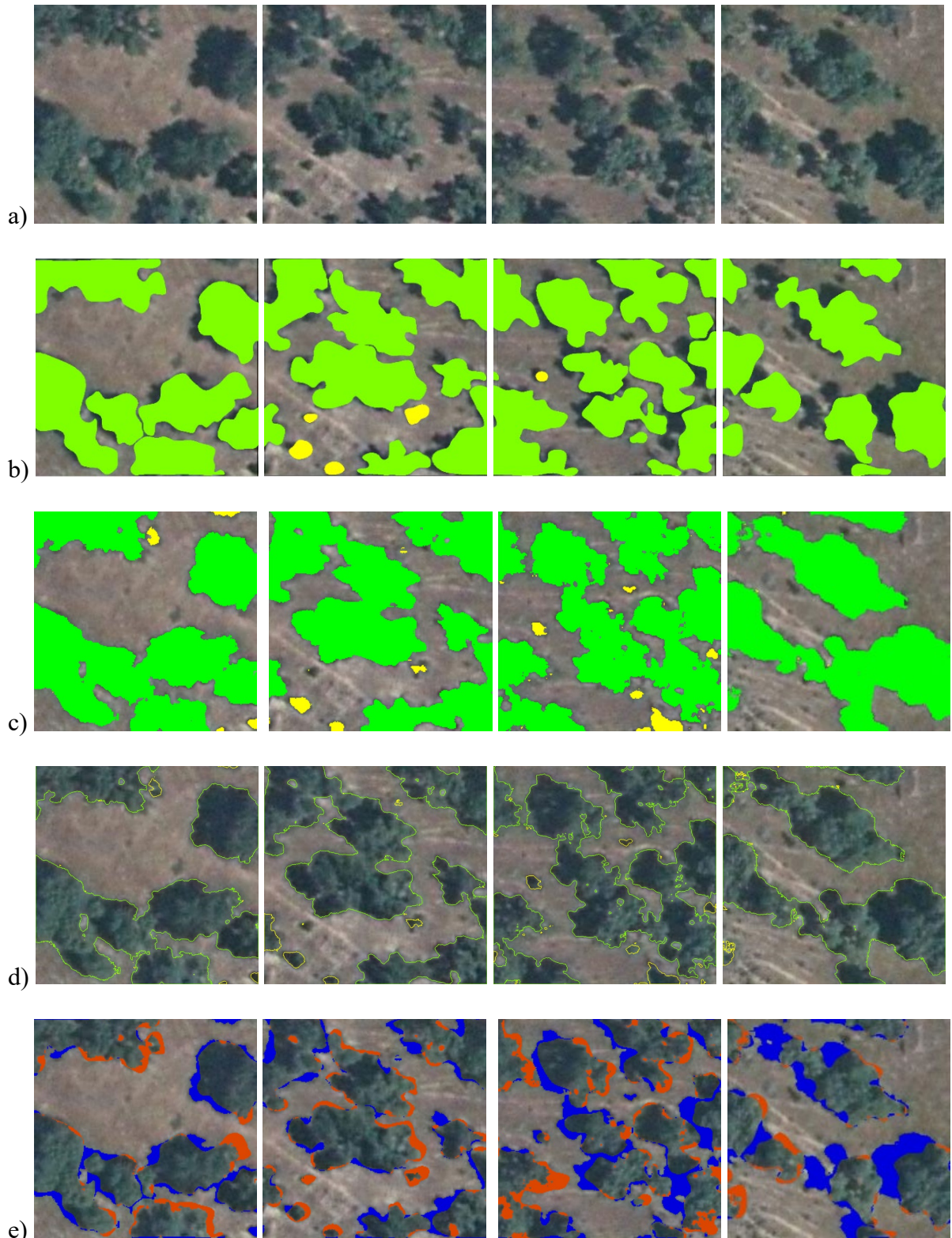

Figure 13: Detection of the crown of trees and acorns. a) Original images Im_70 to Im_73, b) manual segmentation, c) automatic segmentation, d) Detection of the contour of the tree crown and acorn shrubbery e) Comparison between automatic segmentation versus manual segmentation, FPR in blue and FNR in orange.

As shown in Figure 13e, false positives have to do mainly with shadows because in the automatic segmentation they are not discriminated, although in manual segmentation it is not very clear where the threshold between treetops and shadows is established. From the analysis of the manual segmentation of the total orthophotos it is observed that as the experts segmented the orthophotos they eliminated a greater area corresponding to the shadows; this data can be verified from the SAC of the images, as well as the corresponding NU values, that is why in Figure 13e a large number of false positives is observed. On the other hand, false negatives represent the pixels that automatic segmentation detected less than manual segmentation. However, if we analyse Figure 13b, we can see that manual segmentation includes soil pixels and hence these differences.

Figure 14 shows the differences between automatic and manual segmentations according to the quantification of the *FPR* and *FNR* of each of the 38 orthophotos. Except for orthophotos three and five, the values of *FPR* and *FNR* are very similar for the first 15 orthophotos. This similarity is because the group of experts initially considered the shadows as part of the crown of the acorn trees (see Figure 15) and it is from orthophoto 16, where the shadows are already more evident, that they begin to eliminate part of this area. Thus, it is from image 16, except for orthophotos 18 and 21, that the *FPR* is going to rise while the *FNR* is going down, which indicates that not only began to eliminate part of the shadow but also that they learned to better delimit the border between the tree canopy and the ground. This evolution can be seen in Figure 15 for the orthophotos Im_61, Im69, Im_74 and Im92, that is, one of the first images, two intermediate and one of the last.

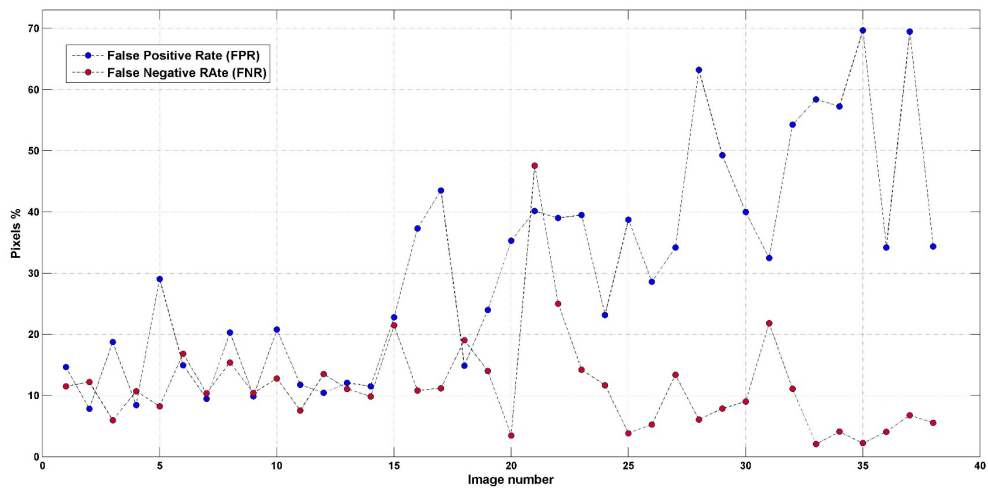

Figure 14: Result of false positives and false negatives in each of the 38 orthophotos.

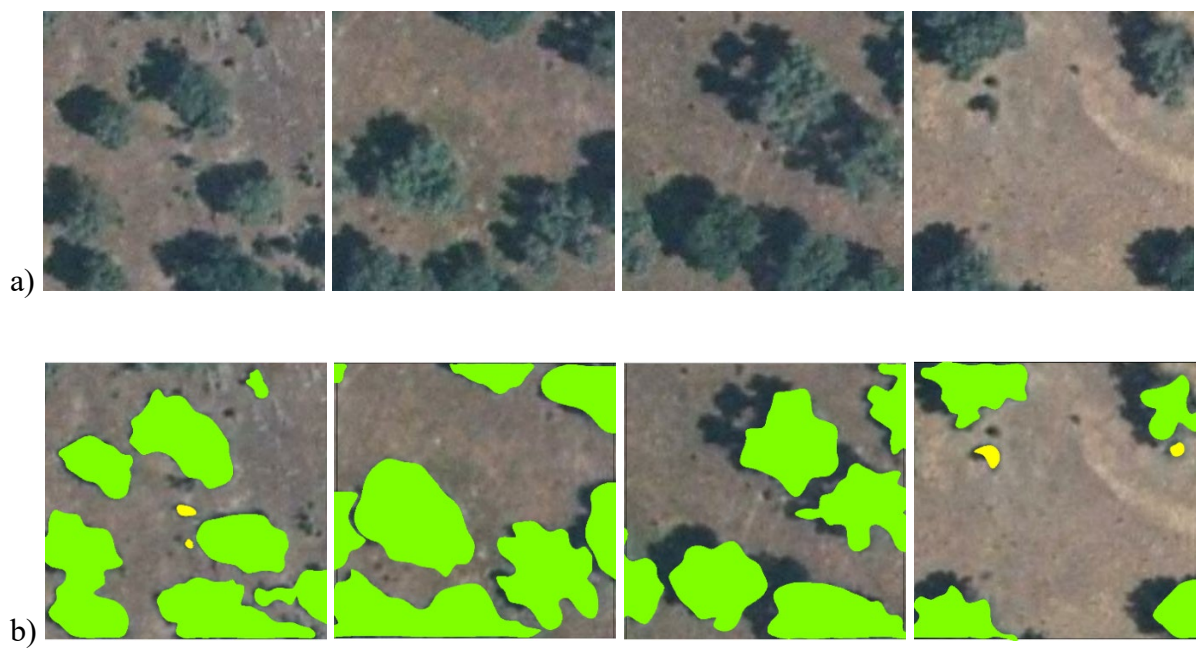

Figure 15: **a)** Original orthophotos: Im_61, Im_69, Im_74 and Im_92. **b)** Manual segmentation of acorn trees.

## Discussion

As we have seen in the results, the detection and quantification of the acorn tree crown area is not an easy task, even for an expert, because the difficulty in some images to differentiate the trees and shrubs from the soil, shadows produced by the trees and the impossibility of clearly setting a threshold that establishes the separation between tree crown and shadows, to the irregularities of the tree crown, to the cavities that could be detected by automatic segmentation but were not reported in the results of manual segmentation. In some areas, the presence of several trees together is evident, although this has not caused as many problems for identification as the presence of shadows. In addition, with automatic segmentation it has also been possible to differentiate acorn trees from shrubs, although in some images only small sections of the tree crown appear because most of them appear in some of the adjacent images and this leads, in some cases, to identify them as shrubs, although due to their shape, size and distribution they could be associated with acorn trees. See Figure 7 for example.

To have an estimate of the homogeneity of the objects found in the 38 orthophotos we used an NU measure whose values, although they are close to zero, are not so small except for the last images. These values occur for both manual segmentation and automatic segmentation, although the latter, except for six cases, has smaller NU values for the first 24 orthophotos, while for the last 14 orthophotos, except for image 31, presents larger NU values. Even if you have this relationship of order in the quantification of homogeneity, the trends are very similar, as can be seen in Figure 11, and this represents, to some extent, the variable difficulty for the correct identification of the crown of trees of acorn and the SAC accordingly.

Once the crowns of the acorn trees were identified in the 38 orthophotos, we proceeded to calculate the SAC for each image, as shown in Figure 12, where we could observe that the values obtained are very close between the segmented images manually and those segmented automatically, particularly for images 1 to 15, 18, 19 and 21. In these images, can be observed a more significant variation of the results based on manual segmentation, which indicates that the experts had a slight initial tendency towards overestimation and underestimation of treetops. In the rest of the images, the SAC is greater for automatic segmentation, but with a strong correlation with manual segmentation, which is a consequence of the beginning of a greater elimination of the shadows by the experts. See Figure 15. Also, and as can be seen in Figure 14 through the FNR and particularly the last six images, the experts also learned to better delineate the contour of the trees with the ground and the end of the graph indicates a greater closeness between both segmentations with respect to tree-soil separation.

To have indicative values on the results obtained, we proceeded to calculate the SACs and the average NUs obtaining the following: a SAC of 30.85% for manual segmentation and a 36.07% for automatic segmentation; see Table 2. Here it is important to note that both percentages have to be corrected depending on the area of the shadows, for manual segmentation to a lesser extent because it makes a better discrimination of the shadows, but it is obvious in some images that the outline has been drawn more by intuition than by a clear separation between tree canopy and shade. In the case of automatic segmentation, there is an obvious overestimation of the SAC because no part of the shadows has been removed. However, even correcting the SAC of the automatic segmentation this is greater than the SAC of the manual segmentation and the maximum admissible stocking density, calculated by interpolation of data from Table 1, which is greater with the automatic segmentation. In addition, if we compare the NU values of Table 2, we can verify that the automatic segmentation has allowed to determine more homogeneous groups and identify the cavities which represent elements in favour of this procedure.

According to the comparative analysis between the automatic segmentation to manual segmentation and quantified through the RPF and FNR, the results are very similar as can be seen at the beginning of the graphs of Figure 14. However, the differences start accentuating from image 16, differences illustrated in Figure 13e, and that represents a better tree-ground segmentation by experts, but also a more significant restriction in the elimination of shadows. The first case is explained by the reduction of the graph of the FNR, while the second case is due to the increase in the graph of the FPR. Thus, and according to these results, we can consider that the real SAC is between the calculated values following the manual procedure and the automatic procedure.

## Conclusions

The correct estimation of the SAC is important for the "acorn" certification of pork in Spain. In this work, we have used an automatic procedure that has allowed us to estimate a SAC of 36.07% and that is greater than the calculated one of 30.85% from the manual segmentation of the images. However, this result also includes the shadows projected on the same trees or the ground so a correction factor must be applied and try to improve this aspect in future work. Also, and according to the results, the above percentages can be taken as the upper and lower limit, respectively, of the real SAC, so it is more reliable to estimate this value to determine the maximum permissible livestock load in a given region. Also, these results are reinforced by the results of the different comparison criteria. The procedure proposed in this paper is then of great relevance to the producers of certified meat and, if it is possible to eliminate the shadow areas in the automatic segmentation, it can play a crucial role to carry out more accurate estimations of the SACs and with of the maximum livestock loads to optimize the use of resources, as well as the production of high-quality meat.

# Conflicts of Interest

The author(s) declare(s) that there is no conflict of interest regarding the publication of this paper.

# Acknowledgments

Finally, the authors would like to thank the National Council for Science and Technology (CONACyT) of Mexico. Part of this work was supported for the incorporation of new teachers UDG-PTC-1461 SEP program.